\documentclass{article}

\usepackage[square,numbers]{natbib}
\usepackage[preprint]{custom_authors}

\usepackage[utf8]{inputenc} 
\usepackage[T1]{fontenc}    
\usepackage{hyperref}       
\usepackage{url}            
\usepackage{booktabs}       
\usepackage{amsfonts}       
\usepackage{nicefrac}       
\usepackage{microtype}      
\usepackage{xcolor}         

\usepackage{caption}
\usepackage{subcaption}
\usepackage{graphicx}
\usepackage{amsmath}

\title{Scaling Laws for Galaxy Images}

\begin{document}

\maketitle

\begin{center}
    \mbox{Mike Walmsley$^{1,2,\star}$}, \mbox{Micah Bowles$^{2}$} \mbox{Anna M.M.\ Scaife$^{2,3}$}, \mbox{Jason Shingirai Makechemu$^{4}$}, \mbox{Alexander J. Gordon$^{5}$}, \mbox{Annette M.~N.~Ferguson$^{5}$}, \mbox{Robert G.~ Mann$^{5}$}, \mbox{James Pearson$^{6}$}, \mbox{Jürgen J. Popp$^{6}$}, \mbox{Jo Bovy$^{7}$}, \mbox{Josh Speagle$^{1,7,8,9}$}, \mbox{Hugh Dickinson$^{6}$}, \mbox{Lucy Fortson$^{10,11}$}, \mbox{Tobias G\'eron$^{1}$}, \mbox{Sandor Kruk$^{12}$}, \mbox{Chris J.\ Lintott$^{13}$}, \mbox{Kameswara Mantha$^{10,11}$}, \mbox{Devina Mohan$^{2}$}, \mbox{David O'Ryan$^{14}$}, \mbox{Inigo V.\ Slijepevic$^{2}$}
\end{center}

\vskip 0.2in

\footnotesize{
$^{1}$Dunlap Institute for Astronomy \& Astrophysics, University of Toronto, Toronto, Canada\\
$^{2}$Department of Physics \& Astronomy, University of Manchester, Manchester, UK\\
$^{3}$The Alan Turing Institute, Euston Road, London, UK\\
$^{4}$Department of Physics, Lancaster University, Bailrigg, Lancaster, UK\\
$^{5}$Institute for Astronomy, University of Edinburgh, Royal Observatory, Edinburgh, UK\\
$^{6}$School of Physical Sciences, The Open University, Milton Keynes, UK\\
$^{7}$Department of Astronomy \& Astrophysics, University of Toronto, Toronto, Canada\\
$^{8}$Department of Statistical Sciences, University of Toronto, Toronto, Canada\\
$^{9}$Data Sciences Institute, University of Toronto, Toronto, Canada\\
$^{10}$School of Physics and Astronomy, University of Minnesota, Minneapolis, Minnesota, USA\\
$^{11}$Minnesota Institute for Astrophysics,  University of Minnesota, Minneapolis, Minnesota, USA\\
$^{12}$European Space Agency, European Space Astronomy Centre (ESAC),  Madrid, Spain\\
$^{13}$Oxford Astrophysics, Department of Physics, University of Oxford, Oxford, UK\\
$^{14}$Centro de Astrobiología (CAB), CSIC-INTA, Madrid, Spain\\
\\
$\star$ E-mail: \texttt{m.walmsley@utoronto.ca}
}

\vskip 0.2in

\begin{abstract}
We present the first systematic investigation of supervised scaling laws outside of an ImageNet-like context -- on images of galaxies.
We use 840k galaxy images and over 100M annotations by Galaxy Zoo volunteers, comparable in scale to Imagenet-1K.
We find that adding annotated galaxy images provides a power law improvement in performance across all architectures and all tasks, while adding trainable parameters is effective only for some (typically more subjectively challenging) tasks. 
We then compare the downstream performance of finetuned models pretrained on either ImageNet-12k alone vs. additionally pretrained on our galaxy images. We achieve an average relative error rate reduction of 31\% across 5 downstream tasks of scientific interest.
Our finetuned models are more label-efficient and, unlike their ImageNet-12k-pretrained equivalents, often achieve linear transfer performance equal to that of end-to-end finetuning. We find relatively modest additional downstream benefits from scaling model size, implying that scaling alone is not sufficient to address our domain gap, and suggest that practitioners with qualitatively different images might benefit more from in-domain adaption followed by targeted downstream labelling.
\end{abstract}

\vskip 0.1in

\section{Introduction}
\label{sec:intro}

Neural scaling laws suggest that larger models trained on more data will perform better. By extrapolating expected performance, they promise a quantitative guide to help practitioners set priorities \cite{hestness_deep_2017}.
The data supporting these laws in supervised computer vision is drawn from experiments with ImageNet or ImageNet-like datasets \cite{hestness_deep_2017,rosenfeld_constructive_2019,sharma_scaling_2022,zhai_scaling_2022}. But ImageNet is a highly curated subset of all possible images. Are scaling laws reliable for practitioners in fields like cell biology, medical imaging, remote sensing, etc., who work with qualitatively different images?

We present the first systematic investigation of scaling laws outside of an ImageNet-like context -- on images of galaxies.
For our domain-specific dataset, we use 840k galaxy images and over 100M annotations by Galaxy Zoo \cite{Masters2019a} online volunteers, comparable in scale to Imagenet-1K \cite{deng_imagenet_2009} and including more than half of all human galaxy labels ever collected. We use these annotations to pretrain our model to jointly solve a diverse set of 88 supervised galaxy image tasks.

We find a power law relationship between labelled domain-specific pretraining data and loss on these pretraining (upstream) tasks; more labelled data smoothly improves performance for all tasks and all architectures. For a given dataset size, larger models perform better but adding parameters eventually leads to harmful overfitting. We set a new upstream state-of-the-art (SOTA) by scaling model size as far as our new and larger dataset allows (approx. 100M parameters). 

Work on scaling laws often reports substantial improvements to few-shot \cite{Radford2019} and finetuning \cite{fan_scaling_2023} performance, including in the context of domain shift \cite{wenzel_assaying_2022} and continual learning \cite{ramasesh_effect_2021}. 
Foundation models \cite{Bommasani2021} formalize the concept of scaling for downstream adaptability. 
Astronomers need models that can adapt to new science tasks and new telescope images.
But pretraining large models is only useful if they learn features relevant to the downstream task. The generalisation gap between pretraining and in-domain images may outweigh any benefits from scaling. 

The medical imaging community, which shares a need for label-efficient adaption to qualitatively different images, `routinely' \cite{wen_rethinking_2021} applies ImageNet-like pretraining with varying results \cite{raghu_transfusion_2019,mustafa_supervised_2021, cherti_effect_2021,matsoukas_what_2022}. Theory \cite{neyshabur_what_2020} and a medical imaging experiment \cite{albarqouni_systematic_2021} suggest additional training on diverse yet in-domain images may be helpful. 
We use our large galaxy image dataset to test the impact of an additional domain adaption step in our context.  

\begin{figure}
    \centering
    \includegraphics{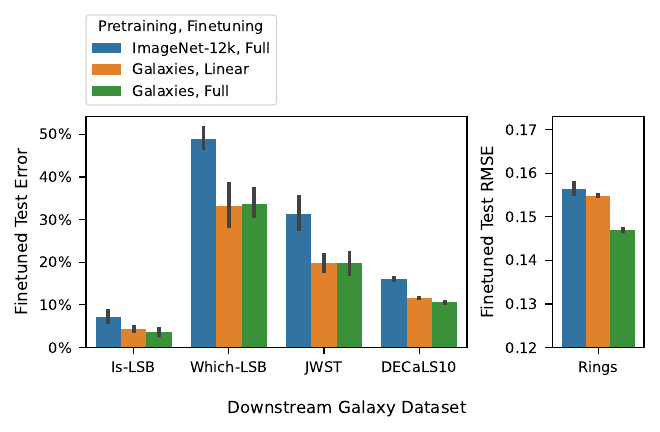}
    \caption{Galaxy images are qualitatively different to common pretraining datasets (e.g.\ ImageNet). Pretraining on labelled galaxy images instead substantially improves performance on diverse downstream galaxy image tasks. See Sec. \ref{sec:downstream_performance} for details.}
    \label{fig:finetuning_summary}
\end{figure}

We compare the downstream performance of models pretrained on either ImageNet-12k\footnote{See \url{https://github.com/rwightman/imagenet-12k} and \cite{rw2019timm}} alone or additionally pretrained on our upstream galaxy tasks, as measured on five downstream galaxy tasks of scientific interest. We find that adding in-domain pretraining on galaxy tasks yields 49\%, 31\%, 37\%, and 34\% relative downstream classification error rate reduction on our Is-LSB, Which-LSB, JWST and DECaLS10 datasets respectively and 6\% relative RMSE reduction on the Rings data-set  (Fig. \ref{fig:finetuning_summary}).
We find relatively modest additional downstream benefits when increasing model size, suggesting that scaling alone is not sufficient to overcome our domain gap, and that practitioners with qualitatively different images might benefit most from in-domain pretraining followed by targeted downstream labelling.

\begin{figure}
\centering

    \begin{subfigure}[b]{0.47\textwidth}
        \includegraphics[width=.8\textwidth]{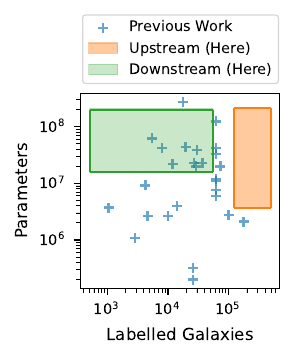}
        \caption{
            Training and parameter scales investigated in this work (upstream and downstream boxes) vs. applied in previous work in our domain (blue points).
        }
        \label{fig:scale_vs_lit}
    \end{subfigure}
    \hfill
     \begin{subfigure}[b]{0.47\textwidth}
    \includegraphics[width=1.\textwidth]{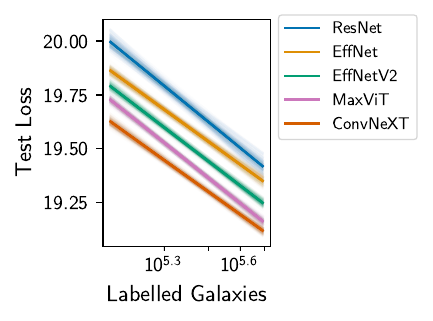}
    \caption{Scaling law fits by architecture family. Posterior median in solid colour. All families show a similar power law improvement with dataset size. 
    }
    \label{fig:scaling_fits}

\end{subfigure}

\caption{Upstream and downstream scales investigated in this work (left) and upstream scaling results (right). We use more labelled galaxy images than any previous work and train larger models than all but one \cite{dagli_astroformer_2023}. Data in Tables \ref{tab:scaling_law_fits} and  \ref{tab:previous_models}.}

\end{figure}

\section{Data}

\subsection{Why are galaxy images different?}

Three key differences between generic and galaxy images are:
\begin{itemize}
    \item \textit{Perspective}. Generic images are framed on a subject and selected with timing and composition, while galaxies are randomly orientated. Leaving our galaxy for a better view is not feasible.
    \item \textit{Softness}. Galaxies are intrinsically amorphous (being comprised of individual stars, dust, etc.). Various effects induced by the telescope optics and by the Earth’s atmosphere also introduce blur.
    \item \textit{Background}. Most pixel values outside the galaxy itself are near zero, affecting image statistics and normalisation.
\end{itemize}

\subsection{Datasets Used}

\begin{figure}
    \centering
    \includegraphics[width=0.9\textwidth]{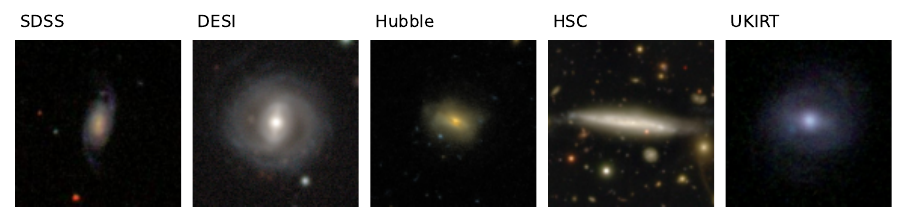}
    \caption{Illustrative galaxy images from our pretraining dataset, split by telescope.}
    \label{fig:examples}
\end{figure}

Astronomers have developed taxonomies describing the visual features (`morphology') of galaxies; for example, our own Milky Way has two (main) spiral arms embedded in a disk, plus a central bar\footnote{See \url{https://science.nasa.gov/resource/the-milky-way-galaxy/}} \cite{vallenari_gaia_2023}. Measuring the appearance of millions of distant galaxies and relating variation in appearance to other physical measurements is key to investigating how galaxies evolve \cite{Masters2019a}.

Our morphology measurements come from Galaxy Zoo, a citizen science project recruiting hundreds of thousands of online volunteers to label images of galaxies. Galaxy Zoo labels have been extensively used as a deep learning benchmark (first by \cite{Dieleman2015}) or as training data for deployed models (e.g.\ for the Euclid space telescope \cite{euclid_collaboration_euclid_2024}). 
These projects use labelled images from a single telescope or, less frequently, apply transfer learning from one telescope to another \cite{DominguezSanchez2019,Tang2019transferLearningRadio,Val2022Self}. 
We draw together 14 years of Galaxy Zoo volunteer annotations (2009-2023) on images from five telescopes to create a combined dataset of 842k galaxy images -- double that of previous work \cite{Walmsley2023desi} and comparable in scale to ImageNet-1k (1.2M images).

Each telescope has different characteristics (wavelength, resolution, etc., see Fig. \ref{fig:examples}) and each labelling campaign asked volunteers different questions about the images. In total we have 88 related galaxy morphology tasks, each of scientific value. We train to answer all tasks jointly with a Dirichlet-Multinomial loss function introduced for this problem. We provide full details on our dataset construction and multi-task loss function in Appendix \ref{sec:upstream_datasets}.

We are deeply grateful to the Galaxy Zoo volunteers, without whom this work would not be possible.

\section{Upstream Scaling Laws}
\label{sec:upstream_scaling_laws}

Fig. \ref{fig:scale_vs_lit} compares the scale investigated here with previous works training galaxy morphology models. Among previous work, there is no relation between training dataset size and selected model size; authors use 2M parameters for 176k images \cite{Walmsley2022decals} to 600M parameters for 17k images \cite{dagli_astroformer_2023}. Most authors train from scratch, a minority use ImageNet pretraining, and a few use galaxy-galaxy transfer learning (e.g.\ \cite{Ghosh2020,DominguezSanchez2019}). 

\begin{figure}
\centering

    \begin{subfigure}{\textwidth}
        \centering
        \includegraphics{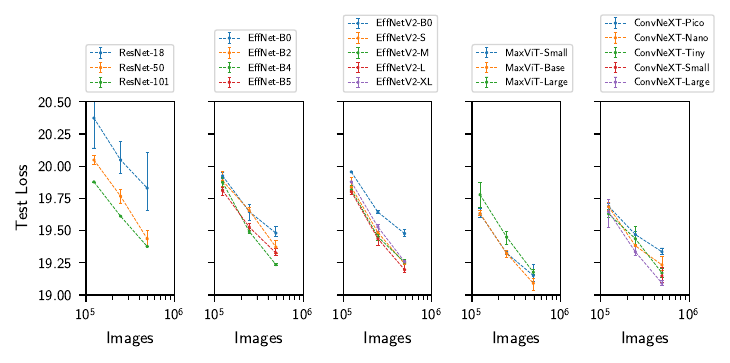}
        \caption{Upstream test loss vs. dataset size, for architecture families (e.g.\ ResNet) and variants (e.g.\ ResNet-50). All families and variants show a power law relationship (of similar exponent) between labelled images and upstream test loss. Adding labelled images leads to far greater improvements in performance than architecture choice.}
        \label{fig:scaling_vs_dataset}
    \end{subfigure}
    
    \begin{subfigure}{\textwidth}
        \centering
        \includegraphics{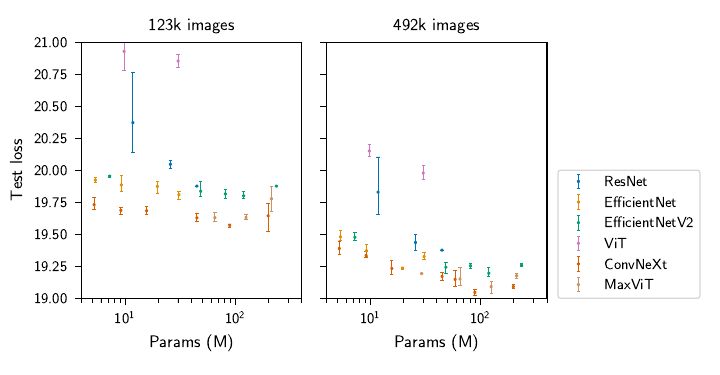}
        \caption{Upstream test loss vs. model size for one-quarter (123k, left) and all (492k, right) galaxy training images. With fewer images, model scaling is only minimally helpful (except for ResNet). With more images, model scaling notably improves performance (up to approx. 100M parameters, using our full dataset). }
        \label{fig:scaling_vs_params}
    \end{subfigure}

    \caption{Upstream performance on galaxy images when scaling dataset size and parameters.}
        
\end{figure}

Figs. \ref{fig:scaling_vs_dataset} and \ref{fig:scaling_vs_params} present our key scaling results. We see a power law between labelled images and test loss
\footnote{We use loss rather than e.g.\ accuracy as our multi-task loss is a smooth measure of statistical success (see \cite{schaeffer_are_2023}). Our loss is a log-likelihood and so we show a linear-log scale.} 
across all architecture families tested (ResNet \cite{He2016a}, EfficientNet \cite{Tan2019a}, EfficientNetV2 \cite{Tan2021}, MaxViT \cite{Tu2022}, ConvNeXT \cite{Liu2022}) and all variants. 
The power-law exponent (i.e. the rate of improvement with more data) is similar (see Appendix \ref{sec:scaling_law_fits} for fits).
This matches \citet{hestness_deep_2017} who claim from ImageNet experiments that architecture changes `only shift the learning curves but do not affect the power-law exponent (though \cite{hestness_deep_2017} only explored ResNet variants for image classification).

When scaling model parameters, we see that large models are most effective only when supported by larger datasets (matching e.g.\ \cite{rosenfeld_scaling_nodate}). We see little-to-no gains when scaling parameters on 123k images (one-quarter of our training dataset) but significant gains on our full training set of 492k images (especially for MaxViT and ConvNeXT, the most recent architectures tested).
Comparing our dataset scale to the training sets in previous works suggests that most galaxy morphology models are likely to be highly overparametrized.

\begin{figure}
     \centering
     \begin{subfigure}[b]{\textwidth}
         \centering
        \includegraphics[width=\textwidth]{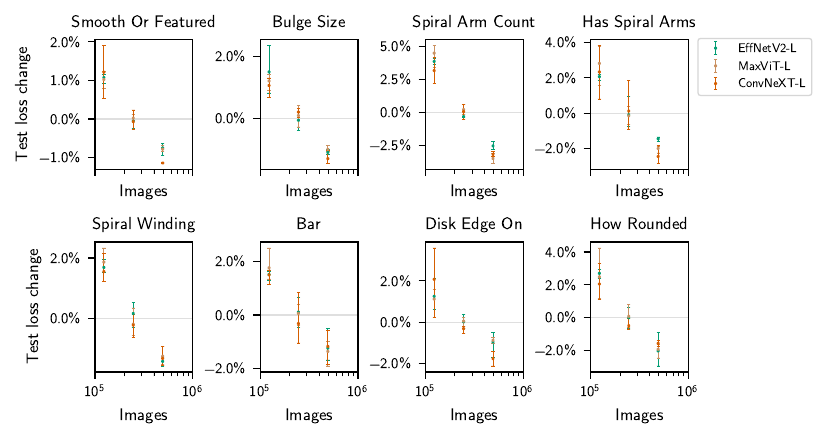}
        \caption{Change in upstream test loss vs. num. labelled images, split by galaxy task, for large models.}
         \label{fig:performance_vs_dataset_by_q}
          \vspace{10mm}
     \end{subfigure}
     \begin{subfigure}[b]{\textwidth}
         \centering
        \includegraphics[width=\textwidth]{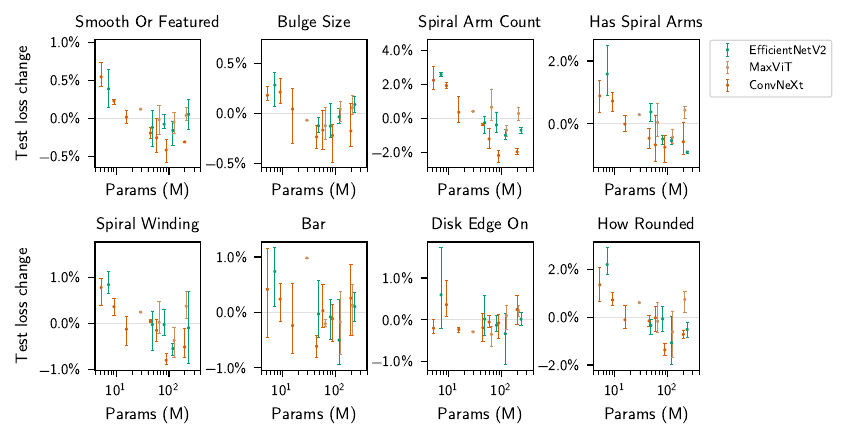}
        \caption{Change in upstream test loss vs. num. parameters, split by galaxy task, when training on our full 492k galaxy image training dataset.}
        \label{fig:performance_vs_params_by_q}
        \vspace{10mm}
     \end{subfigure}
        \caption{Change in upstream test loss vs. num. labelled images (above) or num. parameters (below), split by galaxy Task. More labelled data improves performance as a power law for every individual task. Adding parameters only improves performance at some tasks (e.g.\ `spiral arm count') and has no effect on others (e.g.\ `bar').}
        \label{fig:performance_by_q}
\end{figure}

\begin{figure}
    \centering
    \includegraphics{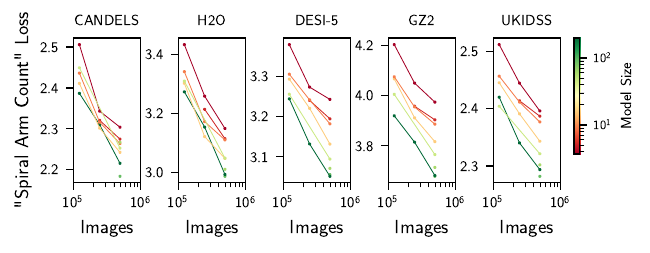}
    \caption{ConvNeXT performance at upstream task `spiral arm count', for each set of telescope images (Hubble and DESI-1/2/8 similar and not shown for clarity). Adding data or parameters both robustly increase performance for this challenging question.}
    \label{fig:convnext_spiral}
\end{figure}

Fig. \ref{fig:performance_by_q} compares the change in test loss for each upstream task\footnote{We aggregate similar tasks across telescopes e.g.\ the test loss shown for `Bar?' is the mean test loss for the `Bar' questions asked for each campaign} as we scale the number of images and model parameters. When adding labelled images, performance consistently improves on every task. 
In contrast, the performance improvement from adding parameters is driven by underlying improvements in some (typically harder) tasks, while some tasks see improvements only from small-to-medium parameter scale (e.g.\ to around 20M parameters) and some see no measurable improvement. We find similar results when comparing similar tasks across different telescopes in Fig. \ref{fig:convnext_spiral}.
Together, this challenges the popular perception from scaling laws that models are either data-limited or parameter-limited. Our models are parameter-limited on \textit{some} tasks while being data-limited on \textit{all} tasks. 

\begin{figure}
    \centering
    \includegraphics{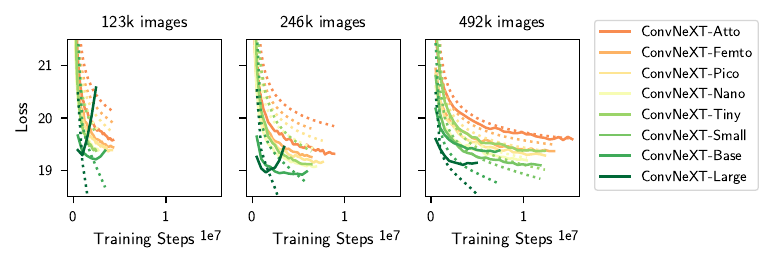}
    \caption{
    Above: ConvNeXT train (dotted) vs. validation (solid) loss curves (averaged over three runs). Larger models rapidly overfit on smaller datasets. Our full training dataset (492k galaxies) supports multi-epoch training of large models, but this dataset scale is not accessible to most astronomy practitioners.}
    \label{fig:overfitting_curves}
\end{figure}

\begin{figure}
    \centering
    \includegraphics{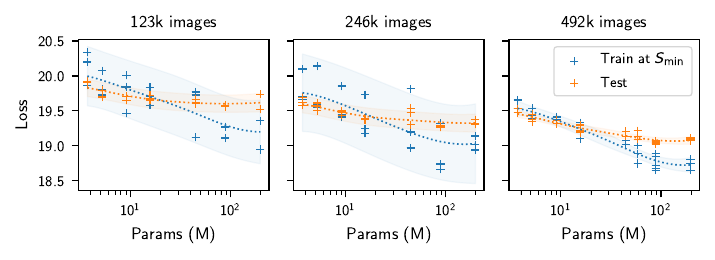}
    \caption{ConvNeXT train vs. test loss upon early stopping ($S_{\text{min}}$). GP-estimated mean and 2$\sigma$ interval shown\protect\footnotemark. Larger ConvNeXT models overfit more at test time.}
    \label{fig:overfitting_vs_params}
\end{figure}

Fig. \ref{fig:overfitting_curves} and Fig. \ref{fig:overfitting_vs_params} investigate the effect of parameter scale on training dynamics, using ConvNeXT variants with 4M to 197M parameters. Adding data dramatically reduces the rate of overfitting, particularly for larger models (Fig. \ref{fig:overfitting_curves}). Selecting evaluation checkpoints based on validation loss broadly avoids catastrophic overfitting at test time (Fig. \ref{fig:overfitting_vs_params}) but increasing parameter count still improves train loss far more rapidly than test loss, suggesting that while a degree of overparameterization is helpful (see e.g.\ \cite{schaeffer_double_2023}), continued parameter-only scaling may be untenable.

\footnotetext{Gaussian process regression with radial basis function and white noise kernels. Length scale fixed at $l=0.6$ to avoid trivial memorisation.}

In summary, we find that adding data provides a consistent power-law improvement across all architectures and tasks, while the improvement from adding parameters is driven by gains on a few (possibly more challenging) tasks. Our quantitative results suggest that most published galaxy morphology models are likely overparametrized. Our larger dataset allows for meaningful scaling up to approx. 100M parameters, outperforming the most closely-related previous work (\cite{Walmsley2023desi}, equivalent to the EfficientNetB0 model in Fig.\ref{fig:scaling_vs_params}), but we suggest that additional parameter scaling may have diminishing returns given the fundamental limits on collecting labelled data.

\section{Downstream Performance}
\label{sec:downstream_performance}

Most astronomers do not have access to annotation capacity on the scale of Galaxy Zoo (over 100k volunteers) and instead rely on labelling images themselves or in small teams. 
Labelled datasets are therefore typically far smaller than that explored here (see Fig. \ref{fig:scale_vs_lit}).
This motivates our search for models that can efficiently adapt to new downstream tasks.

We compare the finetuned performance of models trained from scratch, pretrained on ImageNet-12k alone, or on ImageNet-12k and then additionally on our (upstream) galaxy images. We use our best-performing architecture family (ConvNeXT)\footnote{We also experiment with training EfficientNet-B0 from scratch and chose the higher-performing architecture for each plot (EfficientNet-B0 for `Is LSB' and Galaxy10 DECaLS, and ConvNeXT otherwise).
}. We provide full downstream dataset details in Appendix \ref{sec:downstream_datasets}.

Jointly training on previously-collected labels as an adaption step substantially improves downstream performance at all tasks.
Our adapted models are far more label-efficient; the performance gain is most significant when downstream labels are scarce. ImageNet itself notably outperforms training from scratch. Taken together, our results suggest that ImageNet pretraining is helpful but adaption is vital.

\begin{figure}
    \centering
    \includegraphics[width=0.8\textwidth]{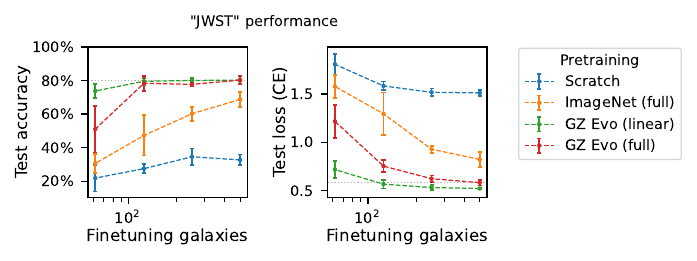}
    \includegraphics[width=0.8\textwidth]{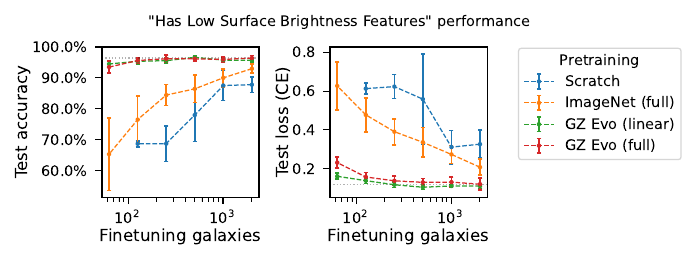}
    \includegraphics[width=0.8\textwidth]{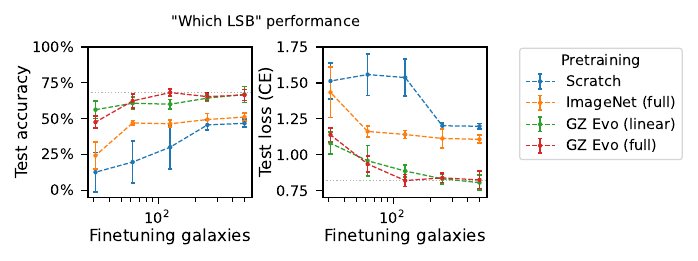}
    \includegraphics[width=0.8\textwidth]{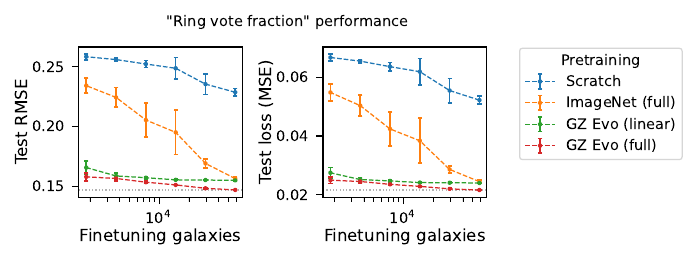}
    \includegraphics[width=0.8\textwidth]{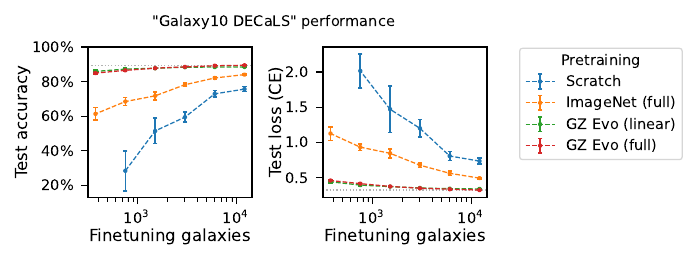}
    \caption{Downstream performance at classifying JWST galaxies (first row) finding (second row) and characterizing (third row) low-surface-brightness features, and regressing ring galaxy vote fractions (fourth row). See Appendix \ref{sec:downstream_datasets} for downstream dataset details.}
    \label{fig:finetuned_inc_scratch}
\end{figure}

We generally achieve equal or close-to-equal performance between linear transfer learning and full fine-tuning. When labels are most scarce (approx. 100-1000 images), linear transfer learning can narrowly (but statistically significantly) outperform full fine-tuning, which we interpret as caused by practical difficulties in optimization. We outperform ImageNet full finetuning with only adaptation and linear transfer learning.

\begin{figure}
    \centering
    \includegraphics[width=0.75\textwidth]{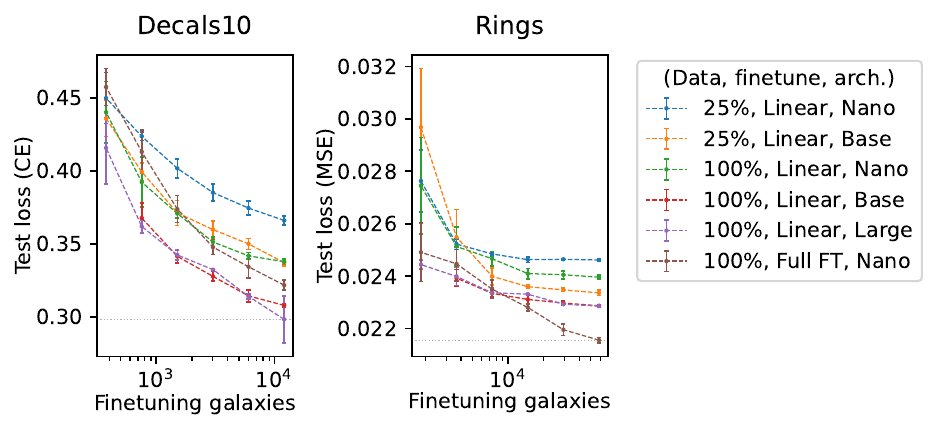}
    \caption{Downstream performance at Galaxy10 DECaLS (left) and GZ Rings (right) depending on upstream scaling. Upstream scaling varies by upstream dataset size (25\% or 100\%) and model size (ConvNeXT-Nano, Base, Large)}
    \label{fig:downstream_with_scaling}
\end{figure}

We find relatively modest additional benefits from scaling up our adaption dataset or our ConvNeXT models, when compared with training from scratch or only on ImageNet-12k (Fig. \ref{fig:downstream_with_scaling}). For both\footnote{We do not present results for the other (smaller) downstream datasets because the train/validation split variance is larger than the downstream performance benefit from upstream scaling.} Galaxy10 DECaLS and GZ Rings, upstream scaling is helpful for downstream performance. The best linear models are the largest variants (ConvNext-Base and ConvNext-Large) trained on the full (100\%, 492k) upstream galaxy dataset, followed by smaller variants (ConvNeXT-Nano) or variants trained on fewer (25\%, 123k) upstream galaxies. We suggest that, for our domain, practitioners should prioritise collecting a diverse and reasonably-large (here, of order $10^6$ examples) adaption dataset over applying the largest possible ImageNet-trained models.

\section{Limitations}
\label{sec:limitations}

We chose to prioritise architectural diversity in our upstream scaling law experiments over highly-tuned pretraining of a specific architecture. 
We selected a simple and reliable pretraining recipe (see Appendix \ref{sec:training_details}) and used it as consistently as possible. The compute cost of our upstream experiments was approximately 14,400 V100-hours.

We present upstream data scaling only across a factor of four; 120k to 490k images. Below this limit, model performance decreases faster than the power law exponent across all architectures, consistent with the `best guessing' transition reported in \cite{hestness_deep_2017}. Testing above this limit is not possible simply because no further human labels exist. We have conclusively tested the data scale likely to be available for supervised galaxy image pretraining in the near term.

This work focuses on the common scenario of pretraining on ImageNet and finetuning on new images, and shows that a domain specific supervised adaption dataset substantially improves performance when the downstream images are qualitatively different to those of ImageNet. Other adaption methods are possible. Notably, self-supervised training on web-scale datasets shows remarkable few- and zero-shot generalization performance (e.g.\ \cite{Radford2019,cherti_reproducible_2022}). We hope this paper will provide a comparison point for future work investigating self-supervised alternatives \cite{Stein2021, Val2022Self, lanusse_astroclip_2023} or hybrid approaches \cite{Walmsley2022Towards}.

\section{Conclusion}
\label{sec:conclusion}

We find that scaling laws broadly apply to the non-ImageNet-like domain of galaxy images. Adding upstream data provides a consistent power-law improvement across all architectures and upstream tasks, while the improvement from adding parameters is driven by gains on a few (possibly more challenging) tasks. We set a new upstream SOTA at around 100M parameters and predict that additional (supervised) parameter scaling is unlikely to be effective given the fundamental limits on collecting additional labelled galaxy data. For adapting to new downstream tasks, we find that additional pretraining on galaxy images leads to far more label-efficient models than with ImageNet-only pretraining; we achieve an average relative error reduction of 31\% across five scientifically-meaningful tasks and achieve comparable linear transfer performance to end-to-end finetuning. Empirically, we find that upstream data and parameter scaling provides relatively modest additional downstream performance gains. 

Our combined results support a middle ground between direct training and finetuning foundation models from generic datasets: additional pretraining with previously-labelled datasets from the same domain followed by finetuning to new tasks through downstream label collection. Pretraining on related datasets provides the scale to support large (enough) foundation models which generalise well, while downstream labels provide a greater relative performance improvement when finetuning for a specific task. In the context of galaxies, this would allow Galaxy Zoo volunteers to participate in more diverse tasks and to make a greater contribution to our understanding of how galaxies evolve. 

We share our improved models as version 2 of the \texttt{zoobot} Python package.
Pretrained encoders are available via HuggingFace at \href{https://huggingface.co/mwalmsley/}{https://huggingface.co/mwalmsley/}. Finetuning code, demos, and documentation are available at \href{https://github.com/mwalmsley/zoobot}{https://github.com/mwalmsley/zoobot}. 

\begin{ack}
MW is a Dunlap Fellow. The Dunlap Institute is funded through an endowment established by the David Dunlap family and the University of Toronto.

LF and KM acknowledge partial support from the  National Science Foundation under grants OAC 1835530 and IIS 2006894.

AMS, MB, DM and IVS gratefully acknowledge support from the UK's Alan Turing Institute under grant EP/V030302/1.

JJP acknowledges funding from the Science and Technology Facilities Council (STFC) Grant Code ST/X508640/1.

CJL acknowledges funding from the Alfred P. Sloan foundation.


\end{ack}

{
\small
\bibliography{main}
}


\appendix

\section{Scaling Law Fits}
\label{sec:scaling_law_fits}

Figure \ref{fig:scaling_vs_dataset} showed each architecture family and variant following a similar power law improvement in test loss with increasing labelled data. Specifically, we noted that each family and variant appears well-fit by a linear model with consistent gradient and varying intercepts. We provide a statistical treatment below.

\begin{figure}
     \centering
        \includegraphics[width=0.5\textwidth]{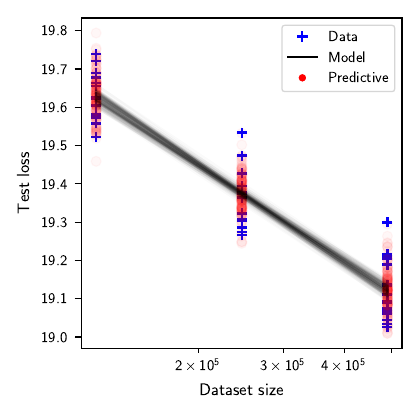}
        \caption{Example scaling law data and fit for an architecture family (here, ConvNeXT). The results of each training run (blue) are fit with a linear model (MCMC samples in black). The quality of the fit is visualised with posterior predictive samples (red).}
         \label{fig:convnext_scaling_fit}
\end{figure}

Figure \ref{fig:convnext_scaling_fit} illustrates our fitting approach. We fit our linear model to data collected from each seeded experiment. We use a Bayesian treatment via ensemble MCMC \cite{Foreman-Mackey2019}. To construct our likelihood, we assume homoskedasticity (i.e. that the random error on the loss for each experiment is drawn from the same distribution) and, under this assumption, find that the random error is well-fit by a normal distribution with $\sigma=0.052$. 

Figure \ref{fig:scaling_fits} (main text, above) compared the resulting fits for each model family. All families show a similar rate of improvement (gradient) with additional labelled data. The choice of family primarily affects the intercept (absolute performance).

The specific parameters of such `scaling laws' are not expected to transfer directly to other problems \cite{hestness_deep_2017} but for completeness we list them in Table \ref{tab:scaling_law_fits}.

\begin{table}[tb]
  \caption{Scaling law parameters for our upstream galaxy morphology dataset (ensemble MCMC fits).}
  \label{tab:scaling_law_fits}
  \centering
  \begin{tabular}{@{}l@{\hskip 11mm}ll@{\hskip 11mm}ll@{}}
    \toprule
    Family & Gradient & (5\%, 95\%) & Intercept & (5\%, 95\%)\\
    \midrule
    ResNet & -0.95 & (-1.07, -0.84) & 24.84 & (24.20, 25.50) \\
    EffNet & -0.85 & (-0.91, -0.79) & 24.19 & (23.85, 24.49) \\
    EffNetV2 & -0.90 & (-0.95, -0.84) & 24.36 & (24.05, 24.65) \\
    MaxViT & -0.94 & (-0.98, -0.89) & 24.49 & (24.24, 24.74) \\
    ConvNeXT & -0.84 & (-0.89, -0.79) & 23.91 & (23.63, 24.18) \\
  \bottomrule
  \end{tabular}
\end{table}

\section{Training Details}
\label{sec:training_details}

\subsection{Upstream Training}

All models are trained with the Adam optimizer and a decoupled weight decay (a.k.a  `adamW') \cite{loshchilov_decoupled_2019} of 0.05. 
All models except for MaxViT and ConvNeXT are trained with a fixed learning rate of $10^{-3}$. For MaxViT and ConvNeXT, we found that training was unstable at this learning rate (particularly for larger variants) and adjusted our recipe to a lower learning rate of $10^{-4}$ and added stochastic depth with a drop probability of 0.4 (based on the ImageNet training recipes used in the model papers).

This adjusted recipe may provide a small test loss advantage to MaxViT and ConvNeXT vs. other models. 
This work is primarily interested in scaling behaviour (which we find to be consistent across model families despite the adjusted training recipe) and not in the absolute performance of each model family.

All models are trained with an effective batch size of 512, achieved by horizontal scaling, gradient accumulation, and synchronized batch-normalisation. For example, for ConvNeXT-Large, we use 8 V100-16GB GPUs (distributing data batches), each with a batch size of 16 and per-GPU gradient accumulation over 4 batches. Efficient distributed I/O via webdatasets was crucial to achieving high GPU utilization. We use mixed precision training throughout. 

We repeat all experiments with at least three random seeds. 
We vary the train/validation split and keep a fixed canonical test set.
All upstream test loss figures display conservative error bars showing the min/max observed test loss. 

Pretrained ImageNet-12k models are sourced from timm \cite{rw2019timm}, including model variants not presented in the original model papers (e.g.\ ConvNeXT-Nano). All models are provided with 224x224x3 pixel images following augmentations typical of the domain (random interpolated rotations, flips, off-center crops).

Models were trained using unallocated capacity on the Digital Research Alliance Canada clusters. 
The GPU time required for our largest models was 153 V100-hours for ConvNext-Large, 170 V100-hours EfficientNetV2-XL, and 236 V100-hours for MaxViT-Large. Across all experiments (4 upstream dataset sizes, 32 model variants, at least 3 seeded runs per combination, plus ad-hoc experiments), our total GPU use was 14,400 V100-hours.

\subsection{Downstream Training}

We replace the upstream head and continue training in one of three configurations. For linear probing i.e. transfer learning, we train the new head with AdamW (learning rate $10^{-4}$, weight decay of 0.1), dropout ($p=0.5$), and a batch size of 32 (on a single V100 GPU). For `full' finetuning, we additionally train the ConvNeXT encoder with a blockwise learning rate decay of 0.25. We use the same protocol for all ConvNexT variants. We test sensitivity to our protocol with hyperparameter searches and find that our results are robust to the choice of protocol parameters (e.g.\ using a larger learning rate does not substantially change performane or alter our conclusions). We again vary the train/validation split and keep a fixed canonical test set.

\section{Upstream Datasets}
\label{sec:upstream_datasets}

\subsection{Galaxy Zoo Labels}

Galaxy Zoo is a crowdsourcing website (\url{www.galaxyzoo.org}) where volunteers are shown a galaxy image and answer a series of questions (e.g.\ `does this galaxy have spiral arms?'). The questions are arranged in a decision tree (formally, in a directed acyclic graph) where the next question asked depends on the answer to the previous question (e.g.\ if `yes, spiral arms', the next question is `how many?'). The labels are recorded as counts per question (e.g.\ `6 of 8 volunteers answered Yes'). Each galaxy is shown to a variable number of volunteers (typically 5 to 40) and the number of answers to each question depend on the earlier choices of the volunteers. 

The decision tree has changed repeatedly over the last 15 years in response to changing scientific priorities, new instruments, and feedback from volunteers \cite{Masters2019a}. Each change (to the tree or to the images shown) is referred to as a new `campaign'. The campaigns used in this work are Galaxy Zoo DESI (397k galaxies, \cite{Walmsley2023desi}) primarily from the Blanco Telescope in Chile,
Galaxy Zoo 2 (188k galaxies, \cite{Willett2013}) from the Sloan Foundation Telescope in New Mexico, Galaxy Zoo Hubble (94k galaxies, \cite{Willett2017a}) and Galaxy Zoo CANDELS (47k images, \cite{Simmons2017}) from the Hubble Space Telescope, Galaxy Zoo UKIDSS (69k galaxies) from the UK Infrared Telescope in Hawaii, and
Galaxy Zoo Cosmic Dawn (45k galaxies) from the Subaru Telescope in Hawaii,
for a total of 842k galaxy images. UKIDSS includes images taken by the UK Infrared Telescope; the survey is described in \cite{Lawrence2007}. Cosmic Dawn includes images taken by the Subaru Telescope as part of the H2O survey; the survey is described at \url{https://project.ifa.hawaii.edu/h20/abstract/}.

Many practitioners have used Galaxy Zoo labels for training deep learning models (see \ref{tab:previous_models} for an incomplete list). These previous works typically train on a single question from a single campaign. Our large-scale upstream dataset is possible because we jointly train on all questions from multiple campaigns. We describe our approach for this in the following section.

\subsection{Multi-Task Learning}
\label{sec:dirichlet_loss}

To jointly train on multiple\footnote{Publication note: an earlier version of this subsection on multi-task loss functions was previously shared as part of the unpublished workshop paper \cite{Walmsley2022Towards} on a different topic (foundation models through hybrid supervised and self-supervised learning).} Galaxy Zoo campaigns, we need to learn from volunteers answering different questions on different images, where any particular image has only a small subset of possible questions answered.

\cite{Walmsley2022decals} introduced the Dirichlet loss function in the context of managing heteroskedastic votes for questions in the same Galaxy Zoo campaign.

\begin{equation}
    \label{multivariate_per_q_likelihood}
    \mathcal{L}_q = \int \text{Multi}(k|\rho, N) \text{Dirichlet}(\rho| \alpha) d\rho
\end{equation}

where, for some target question $q$, $k$ is the (vector) counts of responses (successes) of each answer, $N$ is the total number of responses (trials) to all answers, and $\rho$ is the (vector) probabilities of a volunteer giving each answer. $\rho$ is drawn from $\text{Dirichlet}(\rho|\alpha)$, where the model predicts the Dirichlet concentrations $\alpha$. Multinomial and Dirichlet distributions are conjugates and hence the integral is analytic. Intuitively, this loss corresponds to the odds of observing $k$ heads (votes for an answer) after $N$ coin flips (volunteers asked) assuming a (model-predicted) distribution for the bias of that coin. The Dirichlet-Multinomial distribution allows models to flexibly express uncertainty through wider posteriors and confidence through narrower posteriors.

Assuming answers to different questions are independent, the loss may be applied to multiple questions via the sum
\begin{equation}
    \log \mathcal{L} = \sum_q \mathcal{L}_q(k_q, N_q, f^w_q)
\end{equation}
where, for question $q$, $N_q$ is the total answers, $k_q$ is the observed votes for each answer, and $f^w_q$ is the values of the output units corresponding to those answers (which we interpret as the Dirichlet $\alpha$ parameters in Eqn. \ref{multivariate_per_q_likelihood}).

 To extend to multiple campaigns, we note that this loss naturally handles questions with no answers as $p(a=0|\alpha, N=0)=1$ for all $\alpha$ and hence $\frac{\partial \mathcal{L}}{\partial \alpha} = 0$, meaning unanswered questions do not affect the training gradients. We can therefore construct a multi-campaign vote count vector $K$ where $K_i$ is the votes for answer $i$ and $i$ indexes all answers across all questions \textit{across all campaigns}. For a galaxy labelled in any single campaign, $K_i$ is 0 for any answer $a_i$ to any question not asked in that campaign. Every answer is always predicted but the prediction only affects training if votes for that answer are non-zero. Intuitively, this corresponds to having zero recorded votes to questions not asked. Questions are effectively treated as orthogonal prediction tasks using the same representation. We can therefore jointly train on all answers to all questions to all campaigns.

\section{Downstream Datasets}
\label{sec:downstream_datasets}

\begin{figure}
    \centering
    \includegraphics[width=\textwidth]{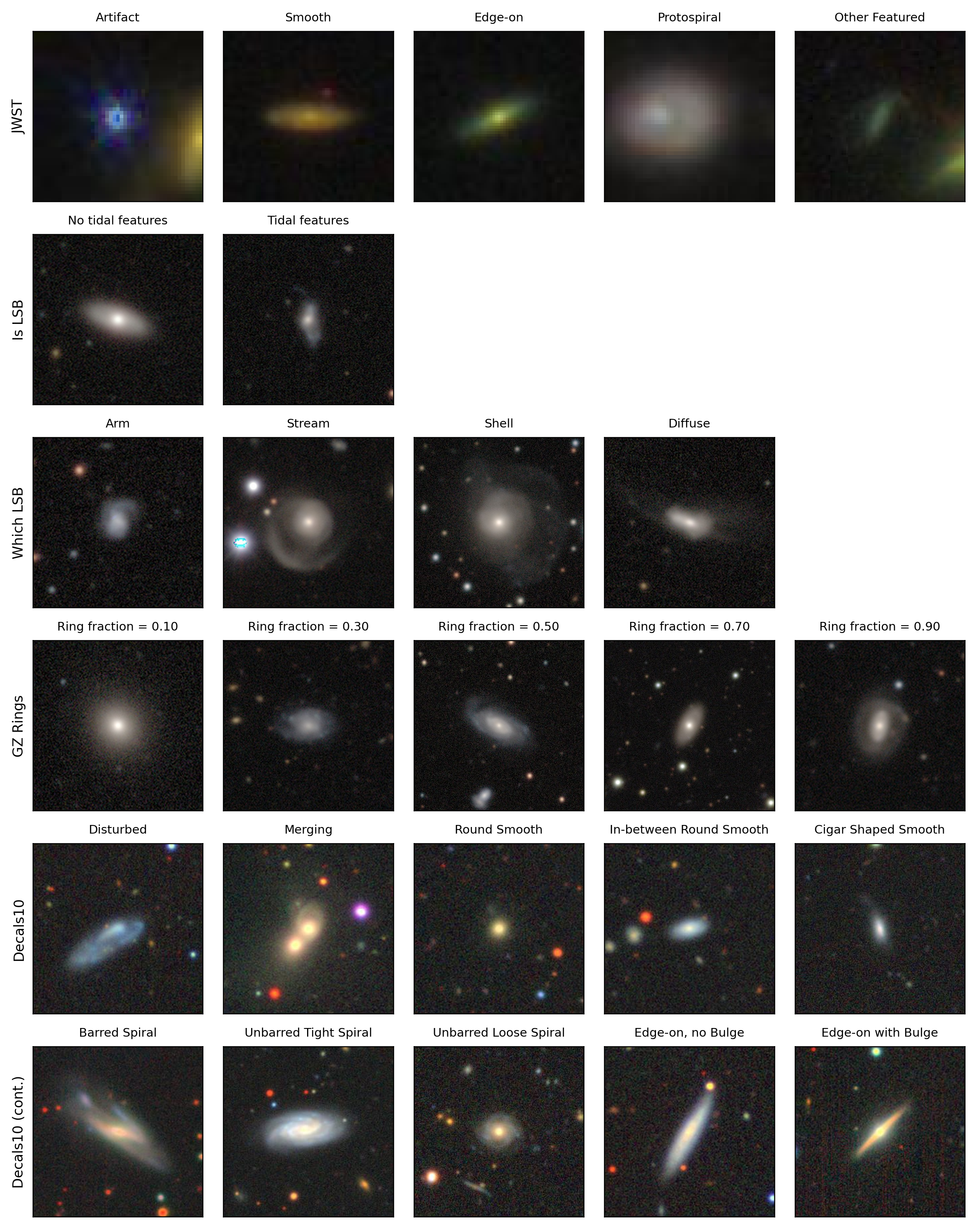}
    \caption{Random examples of labelled galaxies in each downstream dataset. Each dataset is shown by row (two rows for Galaxy10 DECaLS' 10 classes). The downstream label is shown above each galaxy.}
    \label{fig:downstream_datasets}
\end{figure}

Figure \ref{fig:downstream_datasets} shows random examples of the labelled images used in each downstream dataset. We note additional details for each dataset in the subsections below.

\subsection{Classifying Images from the James Webb Space Telescope}

A key practical test of adaptability in astronomy is generalisation to new telescopes and new kinds of galaxies. The James Webb Space Telescope (JWST) offers a view of distant (high redshift) galaxies. The observational characteristics of the telescope and the intrinsic nature of the early-universe galaxies poses a real challenging adaption problem.

For JWST, we (JSM) constructed images using data from the James Webb Space Telescope CEERS and Cosmos-Web surveys. 
All galaxies were selected using ‘SEP’, the Python library for Source Extraction and 
Photometry. 
The F115W and the F150W bands were combined into one image to make the blue (B) 
component of the RGB image, the F277W band formed the G component of the RGB image, 
and the F444W band composed the R part of the image. 

No public morphology labels are available for JWST galaxies. We (MW) then labelled a random 500 of the  1500 most complex (as judged by jpg compression size) images according to the following schema: `artifact' (imaging or other issue), `featureless' (galaxy appears as a featureless blob), `edge-on' (galaxy is visibly an edge-on disk), `protospiral' (galaxy shows signs of spiral structure or other disk-distinguishing features), and `other featured' (galaxy has features that are not spiral-like). We also marked and discarded images as `uncertain' where the true class was not clear. This is our smallest dataset (500 total labels). 

We achieve a 2x error reduction over ImageNet-12k with all 500 labels, and a greater reduction when using fewer labels. Linear evaluation narrowly but statistically-significantly outperforms full finetuning, likely due to our extremely small dataset.

\subsection{Detecting and Characterizing Low Surface Brightness Features}

Galaxies grow partly by accreting one another in violent collisions. These collisions leave faint but visible remnants (`low surface brightness`, LSB, `features'). Identifying these remnants allows astronomers to measure the rate of galaxy growth. Characterizing the remnants reveals the details of each collision, but remnants are hard to distinguish automatically from typical galaxy structure; finding LSB features has been challenging for both rule-based algorithms and deep learning approaches (e.g.\ \cite{Pawlik2016, Bottrell2019, Pearson2019Mergers}).

Gordon et. al (in prep.) visually inspected 2000 images from DECaLS \cite{Dey2018} and determined whether each image showed LSB features (`Is LSB') and, if so, assigned an LSB subclass (`Which LSB'). Subclasses are `arm', `stream', `shell', or `diffuse', based on the taxonomy in \cite{Atkinson2013}. The images to inspect were prioritised by a previous deep learning search (the automated measurements published in \cite{Walmsley2022decals}).

We achieve excellent performance (95\% with 300 labels, 97\% with 2000) vs. ImageNet pretraining (92\% with 2000). This absolute performance compares well with previous literature (typically 80\%-90\% accurate) though a direct comparison is not possible. 
Adapted finetuning on LSB visually-challenging subclasses achieved lower accuracies (approx. 70\%) but improved on ImageNet finetuning by a factor of 2. 

\subsection{Regression on Ambiguous Ring Galaxies}

As the name suggests, ringed galaxies show a circular feature which may either be embedded within or around the galaxy. For an introduction to galaxy rings, see \cite{Buta2017catalog}. They likely relate to galactic bars or other secular dynamical processes.

We use labels collected via Galaxy Zoo Mobile over two years. Volunteers were shown a galaxy image and asked to swipe according to whether it included a ring. The task is ambiguous; identifying a ring is a subjective judgement that experts may disagree on. We therefore chose to frame this task as a regression problem, where the regression target is the fraction of volunteers who swiped `Yes, ring'.

We find that our adapted models are far more label-efficient than with ImageNet pretraining. This matches our overall results that the fewer downstream labels you have, the more helpful adaption becomes. This dataset is large enough that, using the full dataset, ImageNet pretraining almost matches our adapted models.

\subsection{Literature Comparison on Galaxy10 DECaLS Toy Dataset}

By convention and to allow a comparison with other works, we include a finetuned comparison on the
Galaxy10 DECaLS dataset. Galaxy10 DECaLS is a toy dataset derived from Galaxy Zoo DECaLS \cite{Walmsley2022decals} volunteer labels. Galaxy10 DECaLS groups the volunteer labels (each measuring various galaxy features including spiral arms, bars, etc.) into 10 mutually-exclusive classes (e.g.\ `unbarred tight spiral'). Only galaxies where volunteer agreement was high (with a task-dependent threshold) were included. Galaxy10 DECaLS is not formally published but a full description and code to reproduce are available at \url{https://astronn.readthedocs.io/en/latest/galaxy10.html} and \url{https://github.com/henrysky/Galaxy10}. 

In our controlled setup, adaption far outperforms ImageNet pretraining. Our absolute performance is strong (90\% top-1 accuracy over 10 classes). We again achieve comparable accuracy with both full finetuning and linear evaluation, matching the highest-performing baseline in \cite{dagli_astroformer_2023} (fully-trained EfficientNetV2).

Our absolute performance generally compares well to the literature but is lower than the best recently-reported results (95\%, \cite{dagli_astroformer_2023}). We suggest this is at least partly due to dataset limitations (including duplicates, substantial class confusion, limited label precision, and lack of a canonical train/test split) that make it challenging to precisely assess the accuracy of high-performing models. 

\section{Previous Models}

We presented the model and dataset scales of typical galaxy morphology papers in Fig. \ref{fig:scale_vs_lit}, as a comparison to the model and dataset scales systematically investigated in this work. Table \ref{tab:previous_models} shows the underlying data (collected via a literature review). Model parameters are our lower-bound estimates where not reported in the original paper or otherwise uncertain (marked with *).

\begin{table}
\caption{Scale of Previous Galaxy Morphology Works}
\label{tab:previous_models}
\tiny
\centering
\begin{tabular}{lrrr@{\hskip 11mm}ll}

\toprule
Paper & Params (M) & Training Galaxies & Image Size (px) & Model & Labels \\
\midrule
Dieleman+ 2015 & 42 & 62000 & 45 & Custom CNNs & GZ2-derived "Galaxy Challenge" \\
Huertas-Company+ 2015 & 42 & 7000 & 45 & Dieleman-inspired & Experts \\
Remya and Mohan 2016 & *4 & 62000 & 45 & Custom CNN & GZ2/Galaxy Challenge \\
Abraham+ 2018 & 62 & 5000 & 256 & AlexNet & GZ2 and experts \\
Ackerman+ 2018 & 23 & 3000 & 299 & Exception (pretrained) & GZ1/Darg \\
Dominguez-Sanchez 2018 & 3 & 10000 & 69 & Dieleman-inspired & GZ2 or experts \\
Gonzalez+ 2018 & *44 & 20000 & 448 & YOLO (v1) & GZ2/Galaxy Challenge + authors \\
Katebi+ 2019 & 124 & 62000 & 72 & Custom Capsule Network & GZ2/Galaxy Challenge \\
Khan+ 2019 & 23 & 37000 & 299 & Xception (pretrained) & GZ1 \\
Pearson+ 2019 & *11 & 5000 & 64 & Dieleman-inspired & GZ1/Darg \\
Walmsley+ 2019 & 4 & 1000 & 256 & Custom CNN & Experts \\
Cheng+ 2020 & *4 & 3000 & 50 & Custom CNN & GZ1 \\
Kalvankar+ 2020 & *66 & 23000 & 256 & EfficientNet (B0-B7) & GZ2/Galaxy Challenge \\
Vasquez-Mata+ 2020 & 3 & 5000 & 69 & Dominguez-Sanchez-inspired & Authors \\
Walmsley+ 2020 & 2 & 176000 & 128 & Custom CNN & GZ2 \\
Banerjee+ 2021 & 11 & 62000 & 224 & ResNet18 & GZ2/Galaxy Challenge \\
Cavanagh+ 2021 & *22 & 12000 & 100 & AlexNet/Dieleman/Custom CNN & Experts \\
Cheng+ 2021 & 4 & 3000 & 50 & Custom CNN & GZ1 + authors \\
Lin+ 2021 & 3 & 100000 & 224 & Custom ViT + ResNet50 & GZ2 \\
Ravishankar+ 2021 & 33 & 62000 & 224 & VGG-16-inspired & GZ2/Galaxy Challenge \\
Ciprianovic 2022 & 23 & 27000 & 256 & ResNet50 & GZ2 + GZ DECaLS ("Galaxy10") \\
Kramtsov+ 2022 & *20 & 73000 & 100 & DenseNet201 & GZ2 \\
Walmsley+ 2022 & 5 & 200000 & 224 & EfficientNetB0 & GZ DECaLS (data release) \\
Andrew+ 2023 & *20 & 29000 & 128 & DenseNet201 (+ others) & GZ2/Galaxy Challenge \\
Dagli 2023 & 272 & 18000 & 256 & CoAtNet-inspired & Galaxy10 \\
Li+ 2023 & 8 & 62000 & 80 & Custom Capsule Network & GZ2/Galaxy Challenge \\
Walmsley+ 2023 & 5 & 321000 & 224 & EfficientNetB0 & GZ DESI (data release) \\
Wei+ 2024 & *4 & 28000 & 112 & ResNet-inspired & GZ DECaLS \\
\bottomrule
\end{tabular}
\end{table}

\end{document}